\documentclass{article}
\usepackage{spconf,amsmath,graphicx}

\usepackage{color,xcolor}
\usepackage{graphicx}
\usepackage{algorithm}
\usepackage{algorithmicx}
\usepackage[noend]{algpseudocode}

\usepackage{amssymb}
\usepackage{amsmath}
\usepackage{subfigure}
\usepackage{booktabs}
\usepackage{multirow}
\usepackage{stackengine}
\usepackage{amssymb}
\usepackage{pifont}
\usepackage{caption}
\usepackage{color, colortbl}
\usepackage{url}
\definecolor{Gray}{gray}{0.9}
\definecolor{LightCyan}{rgb}{0.88,1,1}
\definecolor{LightBlue}{rgb}{0.902,0.937,1}

\newcommand{\m}[1]{\mathbf{#1}}

\newcommand{\tabincell}[2]{\begin{tabular}{@{}#1@{}}#2\end{tabular}}
\newcommand{\TODO}[1]{}
\newcommand*\rot{\rotatebox{90}}

\usepackage{diagbox}
\usepackage[utf8]{inputenc}
\usepackage[english]{babel}

\title{Dynamically Pruning SegFormer for Efficient Semantic Segmentation}
\name{Haoli Bai $^{\xi\dagger}$, Hongda Mao $^{\xi} $, Dinesh Nair $^{\xi} $}
\address{$^{\xi} $Amazon.com, Sunnyvale, USA\\
$^{\dagger}$The Chinese University of Hong Kong, Hong Kong, China}

\begin{document}
%
\maketitle
\begin{abstract}
As one of the successful Transformer-based models in computer vision tasks, SegFormer demonstrates superior performance in semantic segmentation. Nevertheless, the high computational cost greatly challenges the deployment of SegFormer on edge devices. 
In this paper, we seek to design a lightweight SegFormer for efficient semantic segmentation. Based on the observation that neurons in SegFormer layers exhibit large variances across different images, we propose a dynamic gated linear layer, which prunes the most uninformative set of neurons based on the input instance.
To improve the dynamically pruned SegFormer, we also introduce two-stage knowledge distillation to transfer the knowledge within the original teacher to the pruned student network.
Experimental results show that our method can significantly reduce the computation overhead of SegFormer without an apparent performance drop. For instance, we can achieve $36.9\%$ mIoU with only $3.3$G FLOPs on ADE20K, saving more than $60\%$ computation with the drop of only $0.5\%$ in mIoU.
\end{abstract}

\section{Introduction}

The recent advances of vision transformers (ViT)\cite{dosovitskiy2021image} have inspired a new series of models in computer vision tasks\cite{touvron2021training,liu2021swin,yuan2021tokens,zheng2021rethinking,xie2021segformer}. 
Among ViT variants,
SegFormer\cite{xie2021segformer} extracts hierarchical representations from the input image with the transformer architecture, which shows superior performance in semantic segmentation over the past convolutional neural networks~(CNNs)\cite{long2015fully,chen2017deeplab,chen2018encoder,zhao2017pyramid,zhao2018icnet}.

Despite the empirical success, the SegFormer architecture still suffers high computational cost that challenges the deployment on low-power devices, such as mobile phones or wristbands.
The challenges are mainly from the increasing width of fully connected layers to extract high-level visual features in the Mix Transformer~(MiT) \cite{xie2021segformer}.
Different from previous CNN-based hierarchical architectures\cite{he2016resnet,he2016resnet16,sandler2018mobilenetv2}, the wide and dense linear layers in the MiT encoder inevitably increase the computation overhead.

In this paper, we aim to design efficient SegFormer architecture by dynamically pruning the redundant neurons in the MiT encoder. 
We find that different neurons in a MiT layer exhibit \textit{large variances} across various input instances. 
Motivated by the observation, it would be promising if we can identify the best set of neurons to explain the current input, while pruning away the rest uninformative ones.
Towards that end, we propose a \textit{dynamic gated linear layer} which computes an instance-wise gate via a lightweight gate predictor. The gate thus selects the best subset of neurons for current input and reduces the computation in the matrix multiplication.
Furthermore, we combine \textit{two-stage knowledge distillation}\cite{jiao2020tinybert} to transfer the knowledge from the original SegFormer (i.e., the teacher) to the dynamically pruned one (i.e., the student). The two-stage distillation minimizes the discrepancy of MiT encoder representations and output logits between the teacher and student model respectively.


Empirical results on ADE20K and CityScape benchmark dataset demonstrate the superiority of our method w.r.t. both performance and efficiency over a number of previous counterparts. For example, the dynamically pruned SegFormer can achieve $36.9\%$ mIoU with only $3.3$G FLOPs, which is far smaller than the original $8.4$G FLOPs with $37.4\%$ mIoU.

\vspace{-2ex}




\section{Preliminaries}


\begin{figure*}[t]
	\subfigure 
	{
	    \includegraphics[height=1.5in]{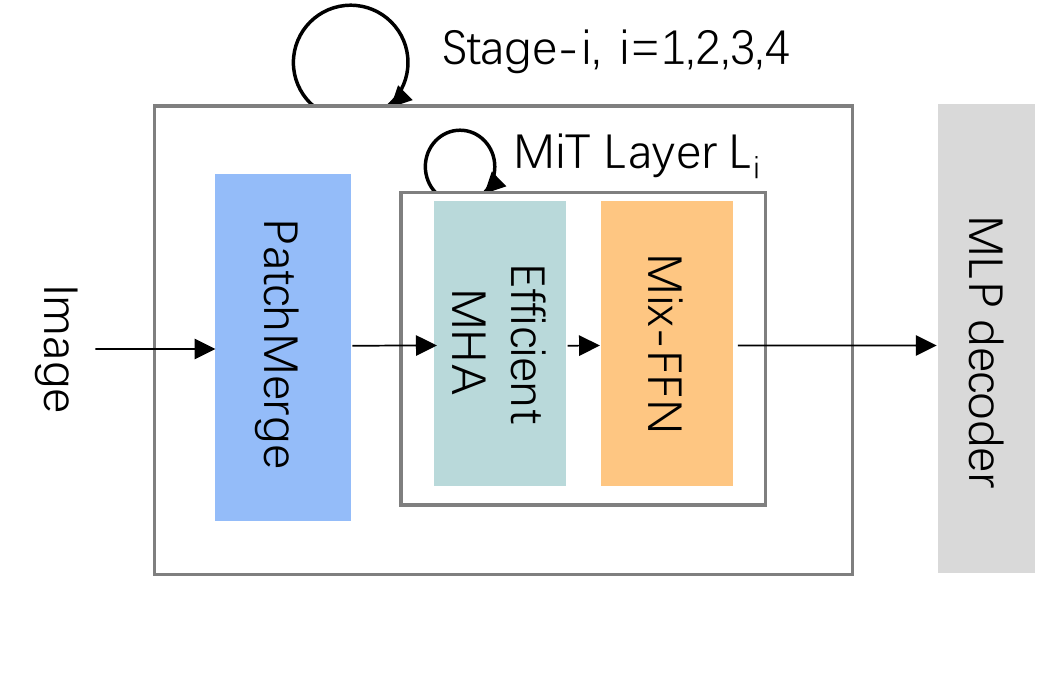}
	    \label{fig:segformer}
	}
	\hfill
	\subfigure
	{
	   \includegraphics[height=1.5in]{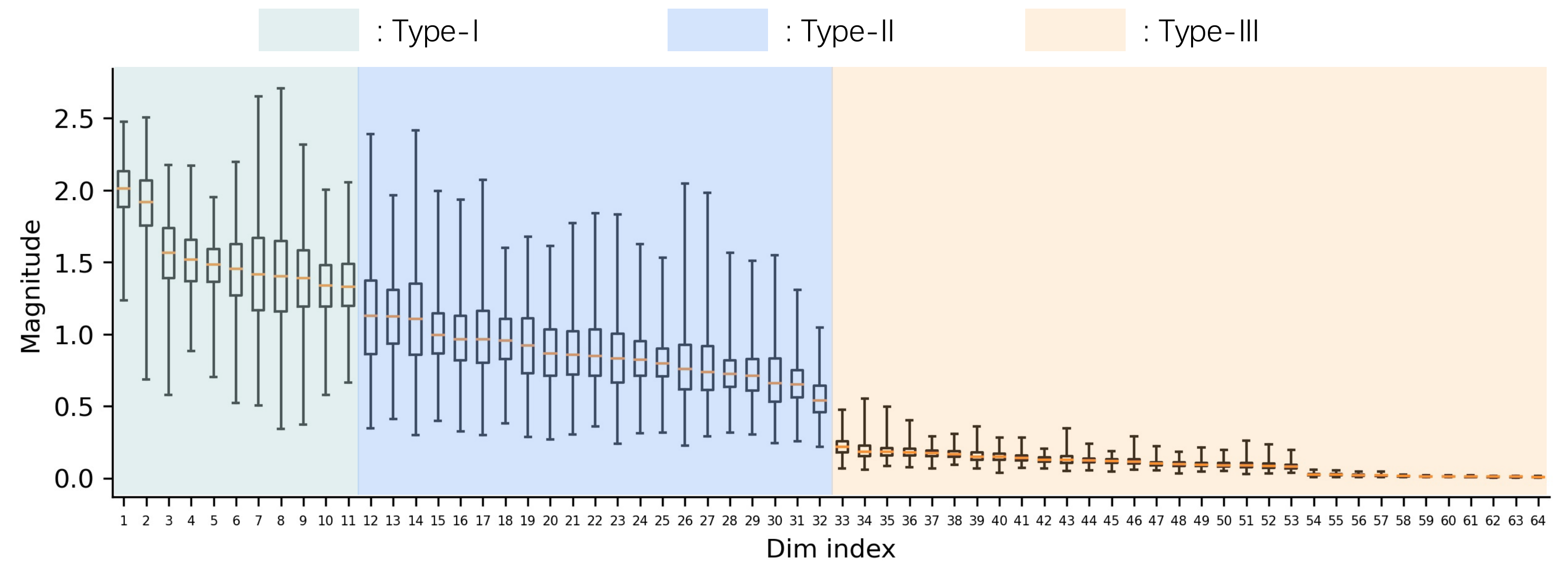}
	   \label{fig:neuron_stats}
	}
	\vspace{-3ex}
    \caption{(\textbf{Left}) shows the architecture of SegFormer, which is composed of four stages. Each stage has multiple MiT layers, where each layer has an MHA module followed by the mix FFN module. The decoder adopts a MLP architecture fo  pixel-wise prediction. (\textbf{Right}) illustrates the neuron variance on the key layer of stage-1.1 in MiT-B0 during inference on ADE20K. The orange lines denote the median, and the rectangles being the lower and upper quartile. The black lines represent the overall range of the neuron.}
    \label{fig:segformer_joint}
\end{figure*}

\vspace{-2ex}
\subsection{The SegFormer Architecture}
SegFormer aims to extract hierarchical feature representations at multiple scales for semantic segmentation.
SegFormer has a Mix Transformer~(MiT) encoder followed by a dense MLP decoder, as shown in the left side of Figure~\ref{fig:segformer_joint}.
The encoder of SegFormer (i.e., MiT) has four different stages. At stage-$i$ of MiT, the input feature map $\bar{\m X}_i \in \mathbb{R}^{C_i \times H \times W}$ is transformed to a patch embedding $\m X_i = \textrm{PatchMerge}(\bar{\m X}_i) \in \mathbb{R}^{N\times C_i}$, where $N$ is the sequence length, and $C_i$ is the hidden dimension for stage-$i$.
The efficient multi-head attention~(MHA) of each transformer layer then computes the query $\m Q = \textrm{FC}(\m X_i) \in \mathbb{R}^{N\times C_i}$, key $\m K  = \textrm{FC}(\m X_i) \in \mathbb{R}^{\bar{N}\times C_i}$ and value $\m V  = \textrm{FC}(\m X_i) \in \mathbb{R}^{\bar{N}\times C_i}$, where $\textrm{FC}(\cdot)$ is the linear layer.
Note that both $\m K$ and $\m V$ have smaller sequence lengths $\bar{N} = N/R $ so as to reduce the time complexity of self-attention to $O(N^2/R)$.
The output of MHA $\m X_{mha}$ thus can be obtained by
\begin{align}
    & \mathbf{A} = \textrm{softmax}(\frac{\m Q\m K^{\top}}{\sqrt{\bar{N}}}) \mathbf{V} \in \mathbb{R}^{N \times C_i} , \\
    & \m X_{mha} = \textrm{LN}\big(\hat{\m X} + \textrm{FC}(\mathbf{A})\big) \in \mathbb{R}^{N \times C_i},
    \vspace{-0.5ex}
\end{align}
where $\textrm{LN}(\cdot)$ denotes the layer-normalization layer.
Then the mix feed-forward network~(Mix-FFN) transforms
$\m X_{mha}$ by:
\begin{equation}
    \m X_{ffn}=\textrm{LN}(\textrm{FC}(\textrm{GeLU}(\textrm{Conv}(\textrm{FC}(\m X_{mha}))) + \m X_{mha}),
    \vspace{-0.5ex}
\end{equation}
where $\textrm{Conv}(\cdot)$ denotes the convolutional operations.
Notably, SegFormer does not employ positional encodings inside the image patches but uses a convolutional layer $\textrm{Conv}(\cdot)$ to leak location information\cite{islam2020much}. 


Finally, the MLP decoder transforms and concatenates the stage-wise feature maps $\{\m X_i\}_{i=1}^4$ together to make the pixel-wise predictions. More details can be found in\cite{xie2021segformer}.




\subsection{Motivation}
\label{sec:motivation}
The hierarchical visual representations in SegFormer also lead to the increasing width in the MiT encoder, which brings more computational burden on low-power devices.


However, we find that each neuron is not always informative across different input instances.
Specifically, we take a trained SegFormer with MiT-B0 encoder to conduct inference over 2,000 instances of the ADE20K benchmark and observe neuron magnitudes across all testing instances. 
We show the box plot of neuron magnitudes of a certain MiT layer on the right side of Figure~\ref{fig:segformer_joint}
The box plot contains the median, quartiles, and min/max values, which can help categorize these neurons into three types:
\textit{Type-I:} large median with a small range, meaning that the neuron is generally informative across different instances; \textit{Type-II:} large median with a large range, meaning that the information of neuron highly depends on the input; \textit{Type-III:} small median with a small range, meaning that these neurons are mostly non-informative regardless of different input.


Given the above observation, it would be promising to reduce the computation inside the MiT encoder by identifying and pruning Type-II and Type-III neurons based on the input.





\section{Methodology}
\label{sec:methods}


\begin{figure}
    \centering
    \includegraphics[width=0.5\textwidth]{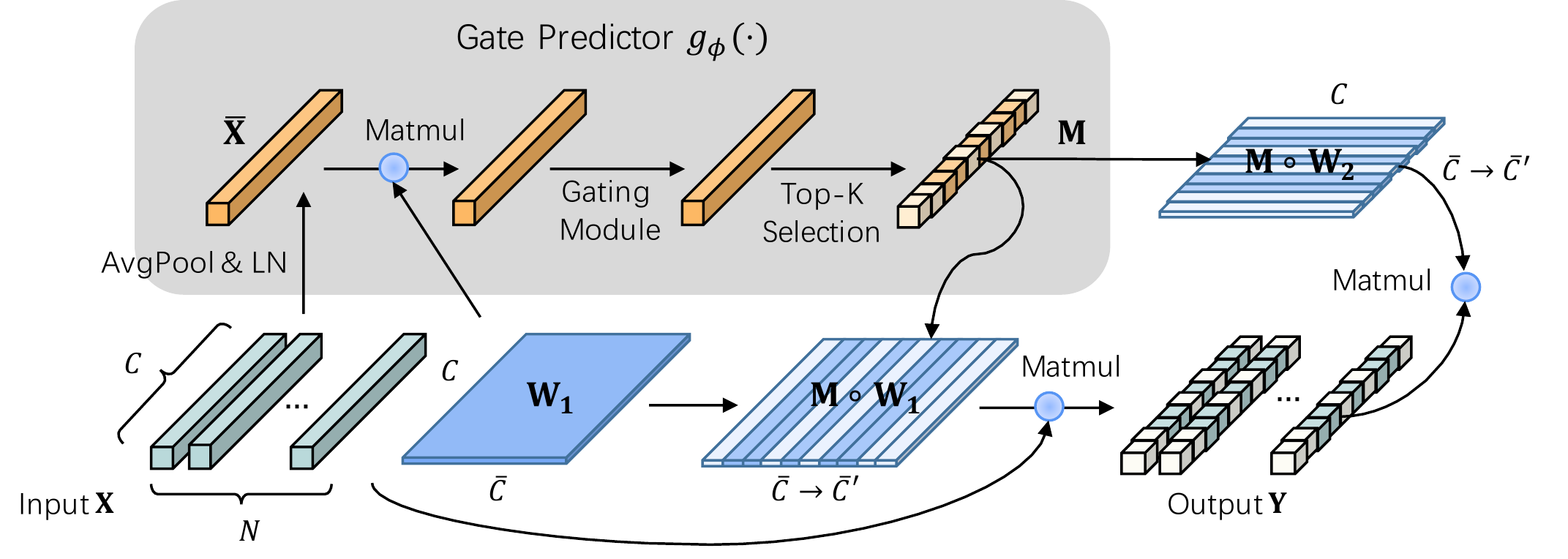}
    \caption{Overview of our dynamic gated linear layer applied to two consecutive $\textrm{FC}(\cdot)$ layers parameterized by $\m W_1$ and $\m W_2$.}
    \label{fig:dgl}
\end{figure}

\subsection{Dynamic Gated Linear Layer}
In order to identify and prune the un-informative neurons based on the input, we propose  \textit{Dynamic Gated Linear}~(DGL) layer, a plug-and-play module to substitute the original linear layer in SegFormer. The DGL structure is shown in Figure~\ref{fig:dgl}.


The workflow of the DGL layer is as follows.
Given input $\m X \in \mathbb{R}^{N\times C}$, the DGL layer computes the gate $\m M = g_{\phi}(\m X) \in \mathbb{R}^{N\times \bar{C}}$ via a gate predictor $g_{\phi}(\cdot)$ parameterized by $\phi$, where is $\bar{C}$ the output dimension. 
The gate $\m M$ can be then applied to mask both the output dimensions of the current linear layer parameter $\m W_1 \in \mathbb{R}^{C\times \bar{C}}$, as well as the input dimensions of the next layer parameter $\m W_2 \in \mathbb{R}^{\bar{C}\times C}$.  
Therefore, the computation can be reduced for two consecutive layers as:
\begin{equation}
    \label{eq:dyanmic}
    \m Y = \m X (\m W_1 \circ \m M) \in \mathbb{R}^{N \times \bar{C}}, \ \ \m Z = \m Y (\m W_2 \circ \m M^{\top}) \in \mathbb{R}^{N \times C}, \nonumber
\end{equation}
where $\circ$ denotes the element-wise product. 

The design of gate predictor $g_{\phi}(\cdot)$ is the key to achieve dynamic pruning.
As the input the gate predictor, we first summarize the sequence by average pooling over the $N$ image patches of input $\m X$, followed by layer-normalization to scale it back to normal ranges, i.e., 
$\hat{\m X} = \textrm{LN}( \textrm{AvgPool}(\m X)) \in \mathbb{R}^{C}$.
Intuitively, aside from the input $\m X$, the parameter $\m W$ should be also incorporated to determine which neuron to prune in the output.
We thus feed both $\hat{\m X}$ and $\m W_1$ into a two-layer $\textrm{MLP}(\cdot)$ activated by $\textrm{ReLU}(\cdot)$ to obtain the mask $\m M$ as
\begin{align}
    \label{eq:gate}
    \m M = \textrm{Top-r}(\m G), \quad \m G = \textrm{MLP}(\hat{\m X} \m W_1) \in \mathbb{R}^{\bar{C}}, \nonumber
\vspace{-0.5ex}
\end{align}
where $\textrm{Top-r}(\cdot)$ keeps the top $r\%$ percentage largest elements in MLP output logits $\m G$, and zero out the rest ones. 
In order to achieve a smooth transition to sparse parameters, we adopt an annealing strategy by gradually increasing the sparsity ratio at the $t$-th step as $r_t = r \min(1, \frac{t}{T})$, where $T$ is total annealing steps.
Note that the $\textrm{Top-r}(\cdot)$ operation inevitably introduces information loss. To remedy this, we also encourage sparsity over the $\m G$, which cab be achieved by $\ell_1$ norm penalty over $\m G$ as the following:
\begin{equation}
\label{eq:sparse_gate}
\vspace{-0.5ex}
\mathcal{L}_{m} = \lambda_m \sum \|\m G\|_1.
\vspace{-0.5ex}
\end{equation}

We deploy the DGL layer to prune $\m Q$, $\m K$ and $\m V$, as well as the intermediate layer of the Mix-FFN of the SegFormer, as a majority of computation lies in the MiT Encoder.
For the MLP decoder, we replace the concatenation by addition over $\{\m X_i\}_{i=1}^4$, so that the computation can be further reduced.


\textbf{Remark} The gate predictor $g_{\phi}(\cdot)$ has only $O(C\bar{C})$ computational complexity, which is negligible compared with the $O(NC\bar{C})$ complexity in the linear layer.
However, all model parameters together with gate predictors should be saved, as every single parameter can be potentially activated depending on the input.
Finally, dynamic pruning has also been previously explored in\cite{gao2018dynamic} for CNNs, where they only pass input $\m X$ to the gate predictor, missing the information inside the parameter $\m W_1$ for the gate prediction.





\subsection{Training with Knowledge Distillation}
The dynamically pruned SegFormer inevitably loses the knowledge from the original model. To bridge the gap incurred by dynamic pruning, we adopt two-stage knowledge distillation\cite{jiao2020tinybert,bai2021binarybert}, which demonstrates superior performance on transformer-based models.
In the first stage, we minimize the mean square error~(MSE) between the student output $\tilde{\m X}_i$ and the teacher output $\m X_i$ at each SegFormer block, i.e.,
\begin{equation}
    \label{eq:stage_1}
    \ell_{1} = \sum_{i=1}^4 \textrm{MSE}(\tilde{\m X}_i, \m X_i).
\vspace{-1ex}    
\end{equation}
Afterwards, with the student logits $\tilde{\m Y}$, we minimize the conventional cross entropy loss (CE) with ground truth data $\m Y^{\star}$ and soft-cross entropy~(SCE) with teacher logits $\m Y$ as:
\begin{equation}
    \label{eq:stage_2}
    \ell_{2} = \textrm{CE}(\tilde{\m Y}, \m Y^{\star}) + \lambda_s \textrm{SCE}(\tilde{\m Y}, \m Y).
\vspace{-1ex}    
\end{equation}
Note that for both two stages, we incorporate the sparsity regularization in Equation~\eqref{eq:sparse_gate}.


\section{Experiments}

\subsection{Experimental Setup}
We empirically verify the proposed approach on ADE20K\cite{zhou2017scene} and Cityscapes\cite{cordts2016cityscapes}, two benchmark dataset for semantic segmentation.
We follow the standard data pre-processing in\cite{xie2021segformer}, i.e., the images of ADE20K and Cityscapes are cropped to $512\times 512$ and $1024\times 1024$ respectively.
We take the batch size as $16$ on ADE20K, and $8$ on Cityscapes, both of which are trained over 8 NVIDIA V100 GPUs.
As we compare both the performance and efficiency of the model, we report the mean Intersection over Union~(mIoU) together with computational FLOPs~(G).

For the main experiments in Section~\ref{sec:main_results}, we present results for both real-time settings and non real-time settings. We adopt MiT-B0 as the encoder backbone for the real-time setting, and MiT-B2 for the non real-time setting.
The dynamic pruning is based on the released model checkpoints~\footnote{\url{https://github.com/NVlabs/SegFormer.}}.
During the training, we take the sparsity regularizer $\lambda_m=0.005$, and soft cross-entropy regularizer $\lambda_s=0.5$ throughout the experiments. The training iterations are set to $160$K, where the first $50\%$ steps are used for sparsity annealing.
We keep the rest configurations consistent with ones used in SegFormer, e.g., with learning rate of $0.00006$ under the ``poly'' LR schedule, and without auxiliary losses and class balance losses.
We name the dynamically pruned SegFormer as DynaSegFormer.
\vspace{-1ex}


\vspace{-2ex}
\subsection{Main Results}
\label{sec:main_results}
\vspace{-1ex}
We present the main results in Table~\ref{tab:main_results}, where $30\%$ and $50\%$ neurons are dynamically pruned in the MiT encoder. It can be found that for both the real-time and non real-time settings on the two benchmarks, our DynaSegFormer can easily outperform previous CNN-based models with significantly fewer computational FLOPs.
For instance, our DynaSegFormer outperforms DeepLabV3+ by $2.9\%$ mIoU with only $5\%$ of its FLOPs.
While the DGL layer increases the size of SegFormer by around $12\%$, it allows to identify and prune the instance-wise redundant neurons given different input.
For example, we can achieve $36.9\%$ mIoU on ADE20K for the real-time setting, which is only $0.5\%$ inferior to SegFormer using only $40\%$ of the original computational FLOPs.

\begin{table*}[t]
    \centering
    \caption{Main results on the validation set of ADE20K and CityScapes. For DynaSegFormer results, $+0.44$ and $+3.35$ are the increased parameter with the proposed DGL layers. }
    \vspace{-1ex}
    \resizebox{0.7\textwidth}{!}{
    \begin{tabular}{llll|lc|lc}
    \hline\hline
    \multirow{2}{*}{} & \multirow{3}{*}{\textbf{\tabincell{c}{Models}}}  &
    \multirow{3}{*}{\textbf{\tabincell{c}{Encoder}}} &
    \multirow{3}{*}{\textbf{\tabincell{c}{Params \\(M) $\downarrow$}}} &
    \multicolumn{2}{c}{\textbf{ADE20K}} &
    \multicolumn{2}{c}{\textbf{Cityscapes}} \\
    
    \cline{5-6} \cline{7-8}
    &&&&
    \tabincell{c}{FLOPs \\ (G)$\downarrow$}  &  \tabincell{c}{mIoU \\ (\%)$\uparrow$} & 
    \tabincell{c}{FLOPs \\(G)$\downarrow$} &  \tabincell{c}{mIoU \\(\%)$\uparrow$} \\

  \hline 
    
    \multirow{7}{*}{\rot{Real-Time}} & 
    FCN\cite{long2015fully} & MobileNet-V2 & $9.8$ & $39.6$ & $19.7$ & $317.1$ & $61.5$ \\
    
    & ICNet\cite{zhao2018icnet} & - & - & - & - & - & $67.7$ \\
    & PSPNet\cite{zhao2017pyramid} & MobileNetV2 & $13.7$ & $52.9$ & $29.6$ & $423.4$ & $70.2$ \\
    & DeepLabV3+\cite{chen2018encoder} & MobileNetV2 & $15.4$ & $69.4$ & $34.0$ & $555.4$ & $75.2$ \\
    & SegFormer\cite{xie2021segformer} & MiT-B0 & $3.8$ & $8.4$ & $37.4$ & $125.5$ & $76.2$ \\ 
    \cline{2-8}
    & \textbf{DynaSegFormer} & MiT-B0 & $3.8_{+0.44}$ & $3.3$ & $36.9$ & $62.7$ & $75.1$ \\
    &  & MiT-B0 & $3.8_{+0.44}$  & $2.7$  & $35.0$ & $46.7$  & $74.3$ \\
    \hline
    \multirow{9}{*}{\rot{Non Real-Time}} & 
    FCN\cite{long2015fully} & ResNet-101 & $68.6$ & $275.7$ & $41.4$ & $2203.3$ & $76.6$ \\
    & EncNet\cite{zhang2018context} & ResNet-101 & $55.1$ & $218.8$ & $44.7$ & $1748.0$ & $6.9$ \\
    & PSPNet\cite{zhao2017pyramid} & ResNet-101 & $68.1$ & $256.4$ & $44.4$ & $2048.9$ & $78.5$ \\
    & CCNet\cite{huang2019ccnet} & ResNet-101 & $68.9$ & $255.1$ & $45.2$ & $2224.8$ & $80.2$ \\    
    & DeepLabV3+\cite{chen2018encoder} & ResNet-101 & $62.7$ & $255.1$ & $44.1$ & $2032.3$ & $80.9$ \\
    & OCRNet\cite{yuan2020object} & HRNet-W48 & $70.5$ & $164.8$ & $45.6$ & $1296.8$ & $81.1$ \\    
    & SegFormer\cite{xie2021segformer} & MiT-B2 & $27.5$ & $62.4$ & $46.5$ & $717.1$ & $81.0$ \\ 
    \cline{2-8}
    & \textbf{DynaSegFormer} & MiT-B2 & $27.5_{+3.35}$ & $18.5$ & $45.4$ & $286.5$ & $80.3$ \\
    &  & MiT-B2 & $27.5_{+3.35}$ & $15.0$ & $44.6$ & $214.2$ & $79.8$ \\
    \hline\hline
    \end{tabular}
    }
    \label{tab:main_results}
\end{table*}
\vspace{-2ex}

\subsection{Discussions}
\vspace{-1ex}
\label{sec:discussion}

For ablation studies, we adopt a MiT-B0 as the encoder under 50\% sparsity for pruning. The model is trained with $40$K iterations for fast verification, and results are present in Table~\ref{tab:ablation}.
We analyze the following aspects:
\textit{\textbf{1) Knowledge Distillation:}}
Comparing with training without distillation (row \#4), two-stage distillation (row \#1) can improve the mIoU by around $1.2\%$. However, distillation with only stage-1 or stage-2 cannot bring such improvement (rows \#2, \#3). The finding is consistent with\cite{jiao2020tinybert,bai2021binarybert} that the technique can better transfer the learned knowledge to the compressed model. 
\textit{\textbf{2) Annealing Sparsity:}} 
It is found generally helpful to anneal the sparsity during the training. 
For instance, it can improve the mIoU by $0.7\%$ for training with distillation (rows \#1, \#5), and $0.5\%$ for training without distillation (row \#4, \#6). The observations match our intuition since a smooth transition of sparsity ratio can better calibrate the parameters throughout the training process.
\textit{ \textbf{3) Dynamic v.s. Static Pruning}}: 
Finally, we verify the advantages of dynamic pruning over static pruning. For static pruning, we follow the widely used magnitude-based pruning\cite{liu2017learning}, where the sparse mask is invariant to different input. It can be found that even with both distillation and sparsity annealing, static pruning is still inferior to dynamic pruning by $2.9\%$ (row \#1, \#7), while the gap is even enlarged to $3.2\%$ when trained without distillation (row \#4, \#8).

\begin{table}[t]
    \centering
    \caption{Ablation studies for pruning types, two-stage knowledge distillation and sparsity annealing.}
    \vspace{-1ex}
    \label{tab:ablation}
    \resizebox{0.45\textwidth}{!}{
    \begin{tabular}{c|lcccc}
    \hline\hline
    \# & \tabincell{c}{Pruning\\Type} & \tabincell{c}{KD\\Stage-1} & \tabincell{c}{KD.\\Stage-2} & \tabincell{c}{Annealing \\Sparsity} & mIoU (\%) \\
    \hline
    1 & Dynamic & \checkmark & \checkmark &
    \checkmark  & $32.1$ \\
    \hline
    2 & Dynamic & \checkmark &  & \checkmark & $30.6$ \\
    3 & Dynamic &  & \checkmark & \checkmark & $30.7$ \\
    4 & Dynamic &  &  & \checkmark & $30.9$ \\
    5 & Dynamic & \checkmark  & \checkmark &  & $31.4$ \\
    6 & Dynamic &  &  &  & $30.4$ \\
    7 & Static & \checkmark & \checkmark & \checkmark & $29.2$ \\
    8 & Static &  &   & \checkmark & $27.7$ \\
    \hline\hline
    \end{tabular}}
    \label{tab:ablation}
\end{table}





\begin{figure}
    \centering
    \includegraphics[width=0.48\textwidth]{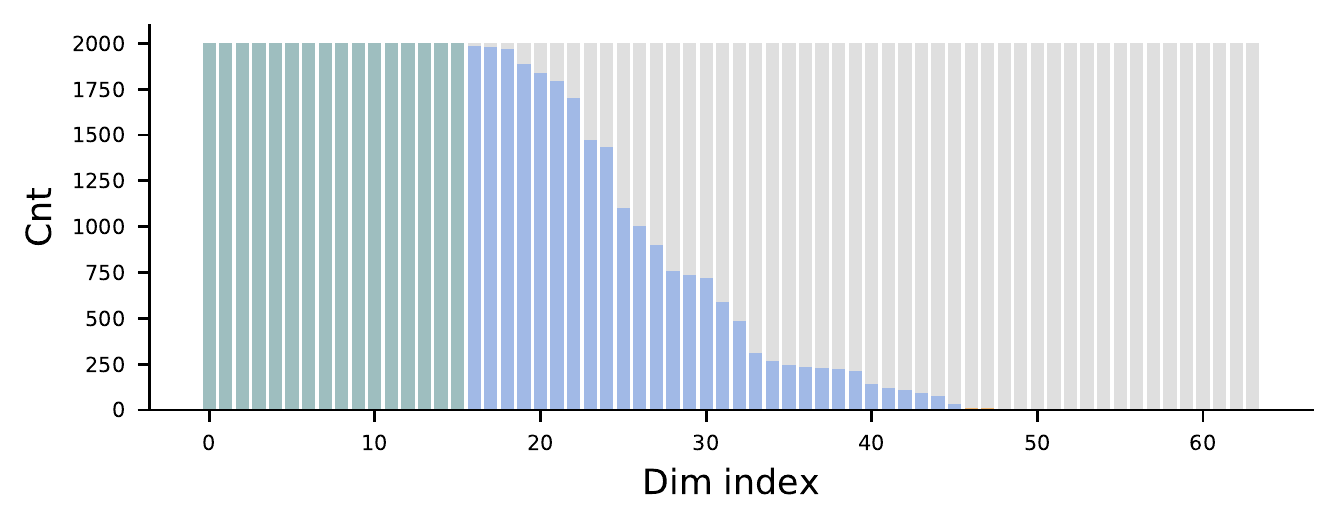}
    \vspace{-2.5ex}
    \caption{Visualization of dynamic gating module on the key layer of stage-1.1 in MiT-B0. The green and blue bars refer to the Type-I and Type-II neurons. The Type-III neurons are absent as they are always not activated during the inference.}
    \vspace{-3ex}
    \label{fig:dyna_gate}
\end{figure}

In the next, we analyze how the learned DGL layer keeps the Type-I neurons, but identifies and prunes the Type-II and Type-III neurons, as mentioned in Section~\ref{sec:motivation}.
We count the number of active gate logits generated by the DGL layer, i.e., $\sum_i \mathbb{I}(G_i>0)$ during the inference, where $\mathbb{I}(\cdot)$ is the indicator function.
From Figure~\ref{fig:dyna_gate}, it can be found that there are around $15$ neurons that are always activated in the $2,000$ testing instances on ADE20K, indicating that they are the Type-I neurons. The Type-II neurons can be selectively activated ranging from $10\sim  1988$ counts, while the Type-III neurons are not visible since they are always inactive throughout all instances. 
Consequently, by identifying the types of neurons, the dynamic scheme selectively prunes neurons to better explain the data under limited computational constraints.

\vspace{-2ex}
\section{Conclusion}
\vspace{-2ex}
In this paper, we propose the dynamic gated linear layer to prune uninformative neurons in SegFormer based on the input instance. 
The dynamic pruning approach can be also combined with two-stage knowledge distillation to further improve the performance.
Empirical results on benchmark datasets demonstrate the effectiveness of our approach.

\bibliographystyle{IEEEbib}
\bibliography{strings}
\end{document}